\documentclass[runningheads]{llncs}
\usepackage[T1]{fontenc}
\usepackage{graphicx}
\usepackage{amsmath,amssymb}
\usepackage{multirow,booktabs,tabularx,array,siunitx}
\usepackage{url}
\usepackage{subfigure}
\usepackage[a4paper,margin=1in]{geometry}

\newcolumntype{Y}{>{\centering\arraybackslash}X}
\usepackage{color}

\makeatother
\begin{document}
\title{Zero-Shot Color Image Manipulation Localization via Noise Residual Artifact Pattern Analysis}
\titlerunning{Zero-Shot Image Manipulation Localization via Artifact Pattern Analysis}

\author{Edgar Gonz\'alez-Fern\'andez\inst{1} \orcidID{0000-0002-1176-2057} %
\authorrunning{E. Gonz\'alez-Fern\'andez}}

\institute{INFOTEC Centro de Investigación e Innovación en Tecnologías de la Información y Comunicación, Aguascalientes, Mexico.
\email{edgar.gonzalezf@infotec.mx}}

\maketitle              % typeset the header of the contribution

\begin{abstract}
Digital cameras embed device-specific artifacts into every acquired image through demosaicing, in-camera post-processing, and lossy compression. These traces constitute a forensic signal that can be exploited to assess image authenticity. Existing passive methods rely predominantly on the green channel of the Bayer residual, discarding the correlated information available in the remaining color channels and typically requiring training data or device enrollment.

This work proposes a zero-shot, training-free blind image manipulation localization pipeline that estimates a reference artifact pattern directly from the noise residual of a single suspect image, without assuming a fixed filter configuration, color layout, or block period.

The pipeline incorporates a principled denoiser selection criterion based on the acquired-to-interpolated noise variance ratio, a block-level correlation analysis against the estimated reference pattern, and a two-component Gaussian Mixture Model scoring stage that produces a pixel-level tampering probability map. An ablation study evaluates the impact of denoiser choice and block size on localization accuracy, and comparisons against state-of-the-art passive methods demonstrate the competitiveness of the proposed zero-shot approach.
\end{abstract}


\section{Introduction}
\label{sec:introduction}
The widespread availability of smartphones, surveillance cameras, drones, and other image-capturing devices has led to a substantial and sustained growth in digital visual content. According to the Reuters Institute Digital News Report 2024, news consumption is increasingly mediated by visual platforms and social networks such as YouTube, Instagram, and TikTok, particularly among younger audiences~\cite{Reuters:2024}. In the United States, more than half of adults report obtaining news from social media platforms, while nearly 40\% of individuals aged 18--29 regularly receive news from online influencers rather than traditional media outlets~\cite{PewSocial:2025,PewInfluencer:2024}. These trends amplify the societal impact of manipulated multimedia content and underscore the need for reliable image authentication mechanisms.

The creation and dissemination of manipulated images is not a recent phenomenon; however, advances in image editing software and, more recently, generative artificial intelligence have made the production of perceptually convincing forgeries accessible to non-expert users~\cite{Verdoliva:2020}. Manipulated images have been employed in political propaganda, misinformation campaigns, financial fraud, and legal disputes, with the potential to influence public opinion before their authenticity can be independently verified~\cite{Farid:2009}. The development of robust forensic techniques capable of assessing the integrity of digital images has therefore become a critical research challenge with direct societal implications.

Existing passive localization methods have predominantly focused on the green channel of the Bayer mosaic, leveraging its higher sampling density, while neglecting the complementary forensic information present in the red and blue channels. Furthermore, most approaches either require a reference image, a device fingerprint library, or supervised training on labeled forgery data, limiting their applicability in real-world scenarios where none of these resources are available.

In this work, we propose a zero-shot passive forensic pipeline for blind image manipulation localization. Operating on a single suspect image without any prior device information or labeled data, the method estimates a reference CFA artifact pattern directly from the full-color noise residual and exploits its spatial coherence through block-level correlation analysis and Gaussian Mixture Model scoring to produce a pixel-level tampering probability map.

% The principal contributions of this work are as follows:

% \begin{itemize}
%     \item A principled denoiser selection criterion based on the
%     acquired-to-interpolated noise variance ratio, enabling
%     data-driven optimization of the residual extraction stage.

%     \item A channel-aware artifact pattern estimation procedure
%     that exploits all color channels of the noise residual and
%     selects the pattern with the strongest periodic structure via
%     DCT-based frequency analysis.

%     \item A zero-shot tampering localization framework combining
%     block-level correlation analysis with a two-component Gaussian
%     Mixture Model that requires no training data, reference images,
%     or device enrollment.

%     \item An ablation study quantifying the effect of denoiser
%     choice and block size on localization accuracy, together with
%     a comparative evaluation against state-of-the-art passive
%     methods on standard forensic benchmarks.
% \end{itemize}

The remainder of this paper is organized as follows. Section~\ref{sec:previous_work} reviews related work in passive image forensics. Section~\ref{sec:acquisition_noise} introduces the image formation model and noise residual framework. The proposed pipeline is described in Section~\ref{sec:methodology}. Experimental results are presented and discussed in Section~\ref{sec:experiments}, and conclusions are drawn in Section~\ref{sec:conclusions}.

\section{Related work}
\label{sec:previous_work}

Noise residuals have become one of the most important sources of information in digital image forensics. A significant body of research exploits Photo-Response Non-Uniformity (PRNU) noise as a device-specific fingerprint for source camera identification and forgery detection \cite{Chen:2008,Cozzolino:2020,Lukas:2006,Marra:2017}. Since PRNU originates from manufacturing imperfections and sensor non-uniformities, images acquired by the same device exhibit similar noise characteristics. Reliable extraction of PRNU traces commonly relies on denoising techniques such as wavelet-based filtering \cite{Donoho:1994,Sendur:2002} and Total Variation methods \cite{Chambolle:2004}, which attempt to suppress image content while preserving acquisition-related artifacts.

Although PRNU-based identification achieves high accuracy, most approaches require a reference fingerprint estimated from a sufficiently large collection of images acquired by the target device \cite{Chierchia:2014,Cozzolino:2020}. This requirement limits their applicability in blind forensic scenarios. To address this issue, Caldelli et al.~\cite{Caldelli:2010} proposed one of the first blind source image clustering methods based on the correlation of noise residuals, demonstrating that images captured by the same device tend to exhibit stronger residual correlations than images from different devices. More recently, several studies have explored statistical properties of noise residuals as compact alternatives to full PRNU estimation, including local variance analysis and residual distribution statistics \cite{Dirik:2009,Ferrara:2012,Retraint:2019,Park:2021}.

A closely related research direction exploits traces left by the Color Filter Array (CFA) and the demosaicing process. Since only one color component is physically measured at each sensor location, the remaining color values must be interpolated, introducing characteristic spatial dependencies. Statistical inconsistencies between acquired and interpolated pixels have been widely employed for forgery detection. Gallagher \cite{Gallagher:2008} demonstrated that variance differences between these pixel populations can reveal CFA configurations, while Kirchner \cite{KirchnerEstimation:2010} exploited re-demosaicing errors to identify interpolation patterns.

Several tampering localization methods have also been derived from the statistical differences between interpolated and acquired pixels. Dirik et al.~\cite{Dirik:2009} proposed a variance-ratio feature for forgery detection, whereas Park et al.~\cite{Park:2021} employed singular value decomposition and prediction residual statistics to identify manipulated regions. More recently, Gonzalez et al.~\cite{Gonzalez:2023} introduced a probabilistic representation of interpolation likelihood, where correlations among local interpolation estimates provide evidence of image manipulation.

The present work lies at the intersection of these research directions. Rather than relying on reference PRNU fingerprints or explicit CFA pattern estimation, it investigates whether compact statistical descriptors extracted from noise residuals can capture acquisition-related traces sufficiently well to support blind image clustering and tampering localization.

\section{Image Acquisition and CFA Artifacts}
\label{sec:acquisition_noise}

This section reviews the concepts underlying the proposed approach. We begin with the image acquisition process and the artifacts it introduces, then formalize the noise residual model that serves as the primary forensic signal throughout this work.

\subsection{Image generation pipeline}
\label{subsec:acquisition}

\begin{figure}
    \centering
    \includegraphics[width=\linewidth]{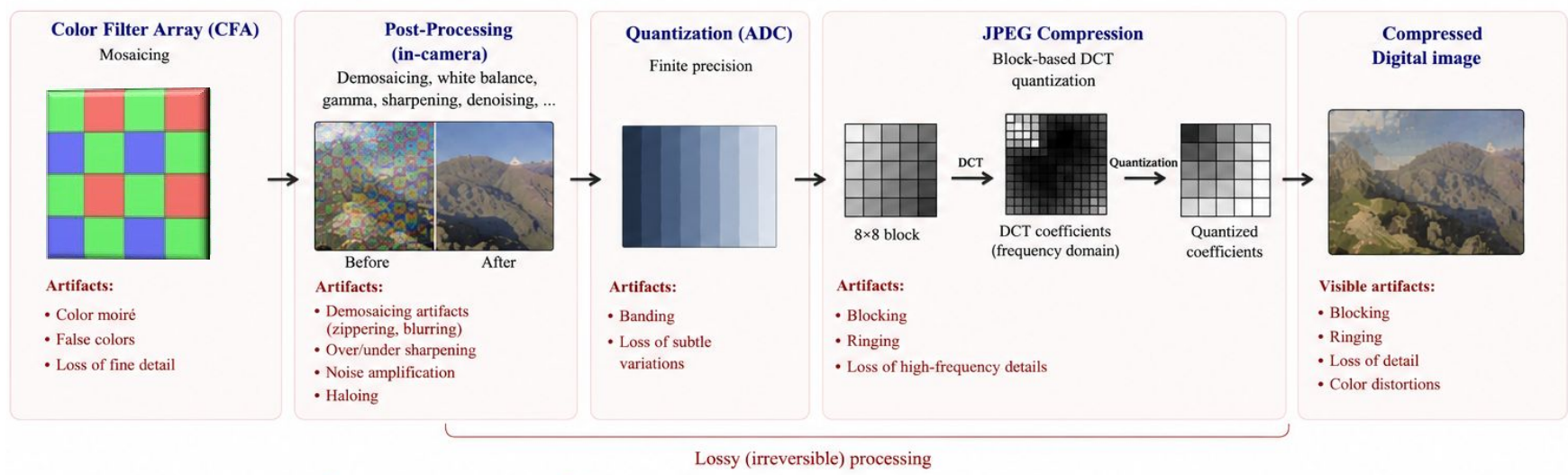}
    \caption{Artifact-inducing stages of the digital image acquisition and compression pipeline. The statistical traces left by lossy, irreversible processing steps are exploited by passive forensic methods for tampering localization.}
    \label{fig:image_pipeline}
\end{figure}

Digital images are the product of a pipeline involving optical lenses, image sensors, color reconstruction, and compression~\cite{Bovik,GonzalezWoods}. At each stage, device-dependent artifacts are embedded into the image, including sensor noise, fixed-pattern noise, and demosaicing traces. Most digital cameras employ a Color Filter Array (CFA) based on the
Bayer pattern, in which each sensor element records only one color channel~(Fig.~\ref{fig:image_pipeline}). The missing color values are recovered
through a demosaicing process that introduces characteristic spatial
dependencies and periodic statistical patterns~\cite{Gallagher:2008,Popescu:2005}.

Image tampering operations such as splicing, copy--move, object removal, and AI-based content synthesis, disrupt these acquisitions traces, inducing statistical inconsistencies between manipulated and authentic regions~\cite{Farid:2009,Verdoliva:2020}. Passive forensic methods exploit precisely these inconsistencies by analyzing noise residuals, CFA interpolation artifacts, and compression signatures, without requiring any reference image or embedded watermark~\cite{Korus:2017,Verdoliva:2020}. CFA artifacts in particular have been widely employed for source identification, tampering detection, and forgery localization~\cite{KirchnerDetection:2010}, and form the basis of the approach developed in this work.

\subsection{Noise Residuals and the Sensor Model}
\label{subsec:noise_model}

The noise residual of an image $\mathbf{I}$ is defined as the difference between the image and a denoised estimate $\mathbf{W} = \mathbf{I} - \mathcal{D}_\theta(\mathbf{I})$, where $\mathcal{D}_\theta$ is a denoising operator with parameters $\theta$. Although traditionally regarded as unwanted signal, such residuals encode information about the acquisition pipeline and have become a primary source of forensic evidence for source attribution and manipulation detection~\cite{Chen:2008,Lukas:2006,Verdoliva:2020}.

Following~\cite{Chen:2008}, a simplified sensor model expresses the
acquired image as
\begin{equation}
    \label{eq:sensor_image}
    \mathbf{I} = \mathbf{I}_0 + \mathbf{I}_0\mathbf{K} + \boldsymbol{\Theta},
\end{equation}
where $\mathbf{I}_0$ is the ideal noise-free scene, $\mathbf{K}$ is a zero-mean multiplicative field representing the Photo-Response Non-Uniformity (PRNU) fingerprint of the sensor, and $\boldsymbol{\Theta}$ aggregates additive noise sources such as dark current, fixed-pattern noise, and shot noise. Given $N$ images $\{\mathbf{I}^{(i)}\}_{i=1}^N$ from the same device, the PRNU
fingerprint can be estimated from their residuals via the maximum likelihood estimator~\cite{Chen:2008}
\begin{equation}
    \label{eq:prnu_estimator}
    \hat{\mathbf{K}} = \frac{\sum_{i=1}^N \mathbf{W}^{(i)}\mathbf{I}^{(i)}}
                            {\sum_{i=1}^N \left(\mathbf{I}^{(i)}\right)^2}.
\end{equation}
Because the residual $\mathbf{W}$ retains CFA interpolation structure when computed with a suitable denoiser, it provides a channel-wise forensic signal sensitive to local manipulation. The empirical selection of $\mathcal{D}_\theta$ is discussed in Section~\ref{sec:denoiser_selection}.

\subsection{Denoising Algorithm Selection}
\label{sec:denoiser_selection}

To extract a meaningful noise residual from a suspect image, the choice of the denoising algorithm and its configuration are consequential: an overly aggressive denoiser suppresses the noise structure that forensic analysis depends on, while an insufficiently strong one leaves scene content in the residual. We therefore evaluate three candidate algorithms across their principal hyperparameters and select the configuration that best preserves the acquired noise signal relative to synthetic interpolated content, as measured by the ratio

\begin{equation}
    \rho(\theta) = \frac{\sigma_A(\theta)}{\sigma_I(\theta)},
    \label{eq:ratio}
\end{equation}

\noindent where $\sigma_A$ and $\sigma_I$ denote the standard deviation of the residual computed over camera-acquired and synthetically interpolated patches, respectively. A higher $\rho$ indicates that the denoiser retains more of the genuine camera noise while attenuating
interpolation artifacts, a desirable behavior for subsequent tamper localization.

Candidate algorithms are taken from the \texttt{skimage.restoration} module~\cite{scikit-image}. For each one, we sweep over the hyperparameters listed below and report the configuration $\theta^*$ that maximizes $\rho$, i.e.
\begin{equation}
    \theta^* = \arg\max_{\theta \in \Theta}\; \rho(\theta),
    \label{eq:theta_star}
\end{equation}
\noindent where $\Theta$ denotes the parameter space of the corresponding algorithm.

\begin{itemize}
\item \textbf{Wavelet-Based Denoising (\texttt{denoise\_wavelet})}. Wavelet shrinkage decomposes the image into a multi-resolution representation and attenuates coefficients below a noise-estimated threshold~\cite{Donoho:1994}. We vary the decomposition depth (wavelet levels) $L \in \{1, 2, 3, 4, 5\}$, the estimated noise standard deviation $\hat{\sigma} \in [0.001, 0.1]$ in steps of $0.001$, and the wavelet family among Haar (\texttt{haar}), Daubechies of orders 4 and 8 (\texttt{db4}, \texttt{db8}), symlets of orders 4 and 8 (\texttt{sym4}, \texttt{sym8}), and the biorthogonal spline wavelet used in the JPEG\,2000 standard (\texttt{bior4.4}).

\item \textbf{Total Variation Denoising} (\texttt{denoise\_tv\_chambolle}). Following Chambolle~\cite{Chambolle:2004}, TV denoising minimizes a weighted sum of data fidelity and total image variation. The sole parameter under study is the regularization weight $\lambda \in \{0.02, 0.05, 0.10, 0.20, 0.50\}$, which controls the degree of smoothing applied to the image.

\item \textbf{Bilateral Filtering} (\texttt{denoise\_bilateral}). Bilateral filtering performs edge-preserving smoothing by weighting neighbors according to both spatial proximity and photometric similarity~\cite{Tomasi:1998}. We explore the window sizes $w \in \{3, 5, 7, 9\}$ (pixels) and the radiometric spread $\sigma_r \in [0.001, 0.1]$ in steps of $0.001$.
\end{itemize}

\begin{table}[!ht]
\renewcommand{\arraystretch}{1.3}
\caption{Best configurations and acquired-to-interpolated standard
deviation ratios $\rho(\theta^*)$ for each candidate denoising
algorithm.}
\label{tab:denoiser_comparison}
\centering
\begin{tabular}{llcc}
\toprule
\textbf{Algorithm} & \textbf{Best parameters} $\theta^*$ & $\boldsymbol{\rho(\theta^*)}$ & \textbf{Selected} \\
\midrule
\texttt{denoise\_wavelet}
    & $L=1$, wavelet$=$bior4.4, $\sigma_{est}$=0.034
    & 1.717 & \\[4pt]
\texttt{denoise\_tv\_chambolle}
    & $\lambda = 0.002$, $\epsilon=0.001$
    & 2.414 & \\[4pt]
\texttt{denoise\_bilateral}
    & $w=3$, $\sigma_r=0.014$
    & 1.235 & \\
\bottomrule
\end{tabular}
\label{tab:denoise_parameter_selection}
\end{table}

\begin{figure}[!ht]
    \centering
    \includegraphics[width=\linewidth]{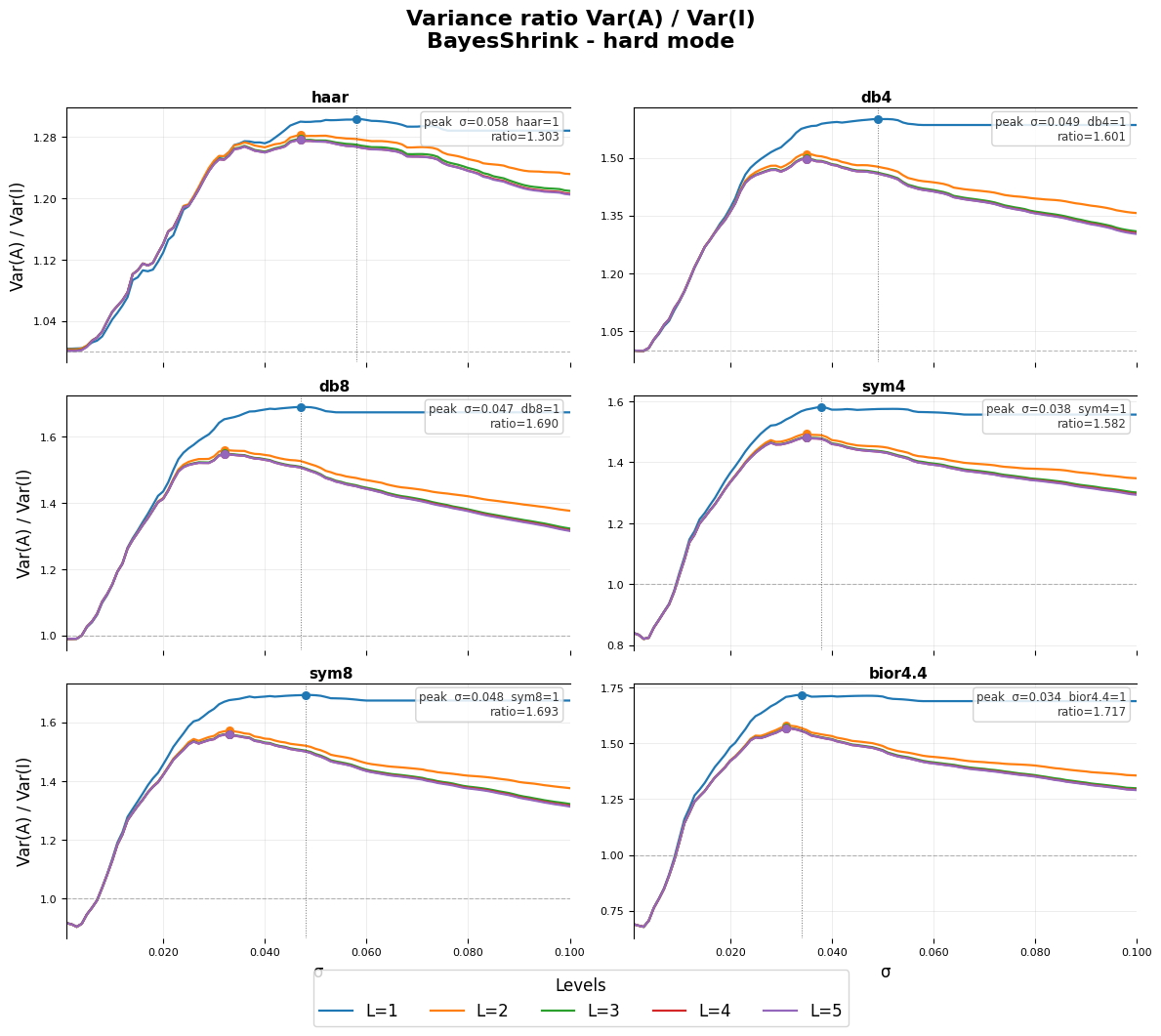}
    \caption{Acquired-to-interpolated variance ratio $\rho(\theta)$ as a function of the estimated noise standard deviation $\hat{\sigma}$
for wavelet-based denoising (\texttt{denoise\_wavelet}), evaluated across decomposition levels $L \in \{1,\ldots,5\}$ and six wavelet
families. In general, a single decomposition level ($L=1$) yields the highest $\rho$, suggesting that shallow decompositions better
preserve the acquired noise structure while suppressing interpolated content.}
    \label{fig:dwt_comparison}
\end{figure}

Figure~\ref{fig:dwt_comparison} shows a comparison of results obtained with wavelet-based denoisers, illustrating the effect of the parameter sweep on the variance ratio criterion. A similar analysis has been performed on TV and bilateral denoising, resulting in the best configurations summarized in Table~\ref{tab:denoiser_comparison}, found for each algorithm together with the corresponding ratio $\rho(\theta^*)$. The algorithm with the highest $\rho(\theta^*)$ overall is carried forward to the tamper map estimation stage (Section~\ref{subsec:pattern_estimation}) and used in the comparisons against state-of-the-art methods (Section~\ref{subsec:results}). 

\section{Feature Extraction}
\label{sec:methodology}

\begin{figure}[!ht]
    \centering
    \includegraphics[width=\linewidth]{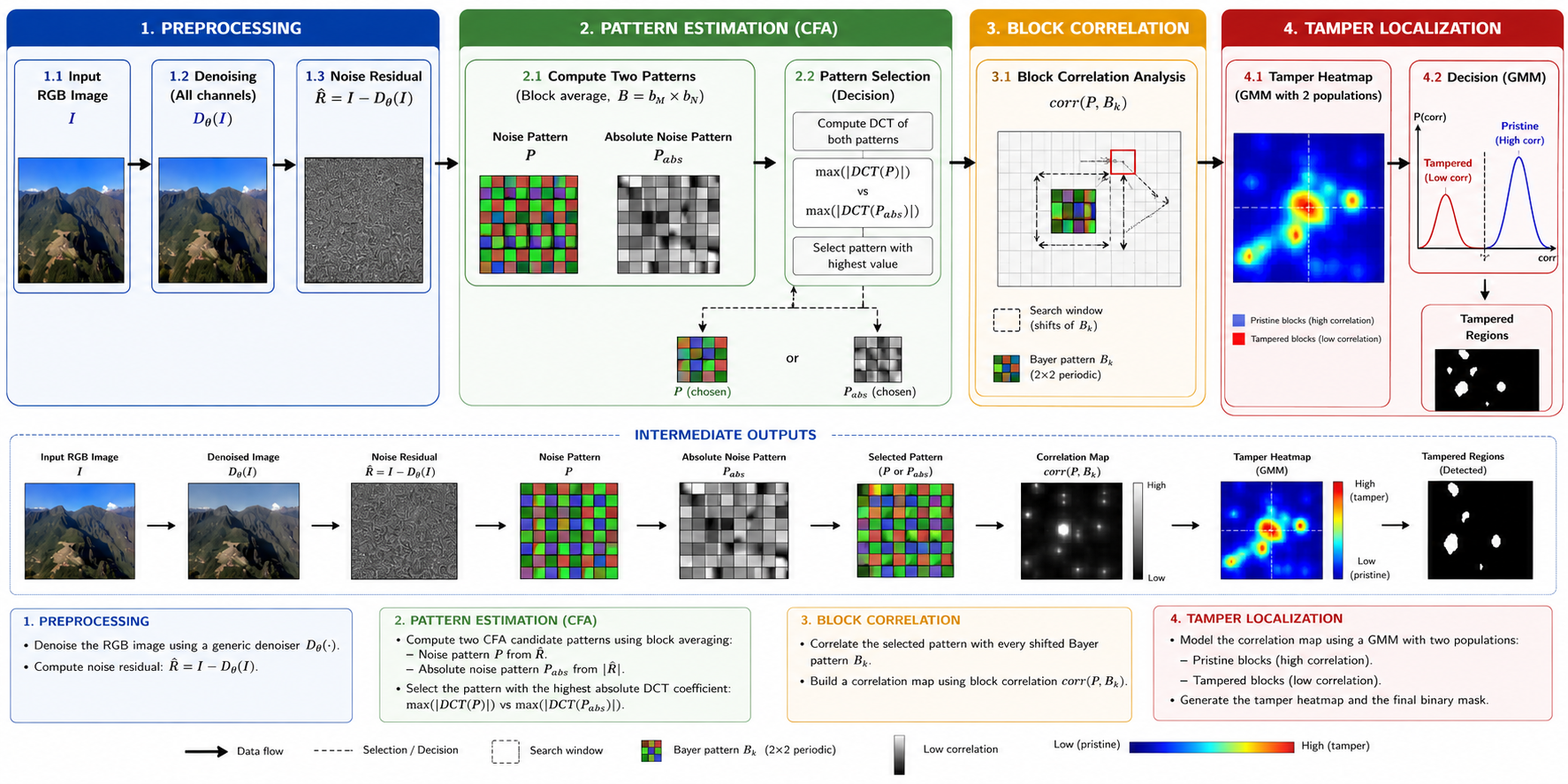}
    \caption{Overview of the proposed zero-shot localization pipeline.}
    \label{fig:methodology}
\end{figure}

The proposed zero-shot blind image manipulation localization pipeline consists of four sequential stages: noise residual extraction, artifact pattern estimation, block-level correlation analysis, and
GMM-based probability scoring. An overview of the full pipeline is illustrated in Figure~\ref{fig:methodology}.

\subsection{Acquisition Artifact Pattern Estimation}
\label{subsec:pattern_estimation}

Given a color image $\mathbf{I} \in \mathbb{R}^{M \times N \times C}$, a noise residual is computed independently for each color channel as
\begin{equation}
    \mathbf{W}_{c} = \mathbf{I}_{c} - \mathcal{D}_{\theta^*}
    \!\left(\mathbf{I}_c\right), \quad c \in \{R, G, B\},
    \label{eq:residual}
\end{equation}
\noindent where $\mathcal{D}_{\theta^*}$ denotes the denoising operator with optimally selected parameters (Section~\ref{sec:denoiser_selection}). The joint residual $\mathbf{W} = (\mathbf{W}_R, \mathbf{W}_G,\mathbf{W}_B)$ contains both genuine camera acquisition noise and residual scene content. To isolate the periodic CFA interpolation fingerprint embedded in $\mathbf{W}$, a reference pattern is constructed by averaging over a non-overlapping partition of the residual into $B \times B$ blocks.

Let $\mathcal{B} = \{\mathbf{B}_k\}_{k=1}^{K}$ denote the set of non-overlapping $B \times B$ blocks partitioning a single-channel residual $\mathbf{W}_c$, where $K = \lfloor M/B \rfloor \lfloor N/B
\rfloor$. Two candidate reference patterns are defined: a signed average
\begin{equation}
    \mathbf{P}^{(1)}_c = \frac{1}{K}\sum_{k=1}^{K} \mathbf{B}_k,
    \label{eq:pattern1}
\end{equation}
and an absolute average
\begin{equation}
    \mathbf{P}^{(2)}_c = \frac{1}{K}\sum_{k=1}^{K} |\mathbf{B}_k|,
    \label{eq:pattern2}
\end{equation}
\noindent where $|\cdot|$ denotes the element-wise absolute value. Both patterns approximate the periodic CFA interpolation fingerprint of a pristine image region: $\mathbf{P}^{(1)}_c$ preserves the sign of the noise fluctuations, while $\mathbf{P}^{(2)}_c$ accentuates the magnitude of the periodic structure irrespective of polarity.

The pattern exhibiting the stronger periodic structure is selected as the reference for the subsequent block correlation stage. Periodicity is assessed in the frequency domain: the
two-dimensional Discrete Cosine Transform (DCT) is applied to each candidate, and the pattern whose non-DC coefficients have the greatest aggregate magnitude is selected,
\begin{equation}
    \mathbf{P}^*_c = \arg\max_{\mathbf{P} \in \{\mathbf{P}^{(1)}_c,\mathbf{P}^{(2)}_c\}}
    \sum_{(u,v) \neq (0,0)}
    \left|\hat{\mathbf{P}}(u,v)\right|,
    \label{eq:pattern_selection}
\end{equation}
\noindent where $\hat{\mathbf{P}}$ denotes the DCT of $\mathbf{P}$. The DC coefficient is excluded as it encodes the mean offset of the pattern rather than its spatial periodicity. The selected pattern
$\mathbf{P}^*_c$ and its corresponding residual form---$\mathbf{W}_c$  if $\mathbf{P}^* = \mathbf{P}^{(1)}_c$, or $|\mathbf{W}_c|$ if $\mathbf{P}^*_c = \mathbf{P}^{(2)}_c$---are carried forward to the
correlation stage, where the block-level feature map is defined as
\begin{equation}
    \mathbf{F}(i,j) = \text{corr}\!\left(\,
        \mathrm{vec}\!\left(\mathbf{P}^*\right),\;
        \mathrm{vec}\!\left(\mathbf{B}^*_{ij}\right)
    \right),
    \label{eq:corr}
\end{equation}
\noindent where $\mathbf{P}^* =  (\mathbf{P}^*_R, \mathbf{P}^*_G,\mathbf{P}^*_B)\in \mathbb{R}^{B \times B \times C}$, $\mathbf{B}^*_{ij} \in \mathbb{R}^{B \times B \times C}$ denotes the
$(i,j)$-th block of the selected color residual, $\mathrm{vec}(\cdot)$ is the vectorization operator that concatenates all color channels into a single vector, and corr denotes the Pearson correlation coefficient. Under the authentic hypothesis, the artifact pattern of a block is statistically
consistent with $\mathbf{P}^*$, yielding $\mathbf{F}(i,j)$ close to unity; a tampered block, whose noise has been replaced or disrupted by the manipulation, produces a correlation near zero.

\subsection{GMM-Based Probability Scoring}
\label{subsec:gmm_scoring}

The correlation values $\{\mathbf{F}(i,j)\}$ are modeled as draws from a two-component Gaussian mixture,
\begin{equation}
    p(f) = \sum_{k=1}^{2} \pi_k \,
    \mathcal{N}\!\left(f;\, \mu_k,\, \sigma_k^2\right),
    \label{eq:gmm}
\end{equation}
\noindent whose parameters $\{\pi_k, \mu_k, \sigma_k^2\}_{k=1}^{2}$ are estimated via Expectation-Maximization. The component with the larger mean $\mu_k$ is designated as the authentic population; the remaining component corresponds to tampered blocks. Each block is then assigned the posterior probability of belonging to the tampered component, yielding a pixel-level soft tampering map $\mathbf{T}$, where each pixel inherits the score of its enclosing block.

A binary localization mask is obtained by thresholding $\mathbf{T}$,
\begin{equation}
    \mathbf{S}(i,j) =
    \begin{cases}
        1, & \text{if } \mathbf{T}(i,j) > \tau, \\
        0, & \text{otherwise},
    \end{cases}
    \label{eq:binary_map}
\end{equation}
\noindent where $\tau$ is determined by Otsu's method~\cite{Otsu:1979}. Pixels labeled $\mathbf{S}(p,q) = 1$ constitute the estimated manipulated region.

\section{Experiments and Results}
\label{sec:experiments}

\begin{figure}
    \centering
    \includegraphics[width=\linewidth]{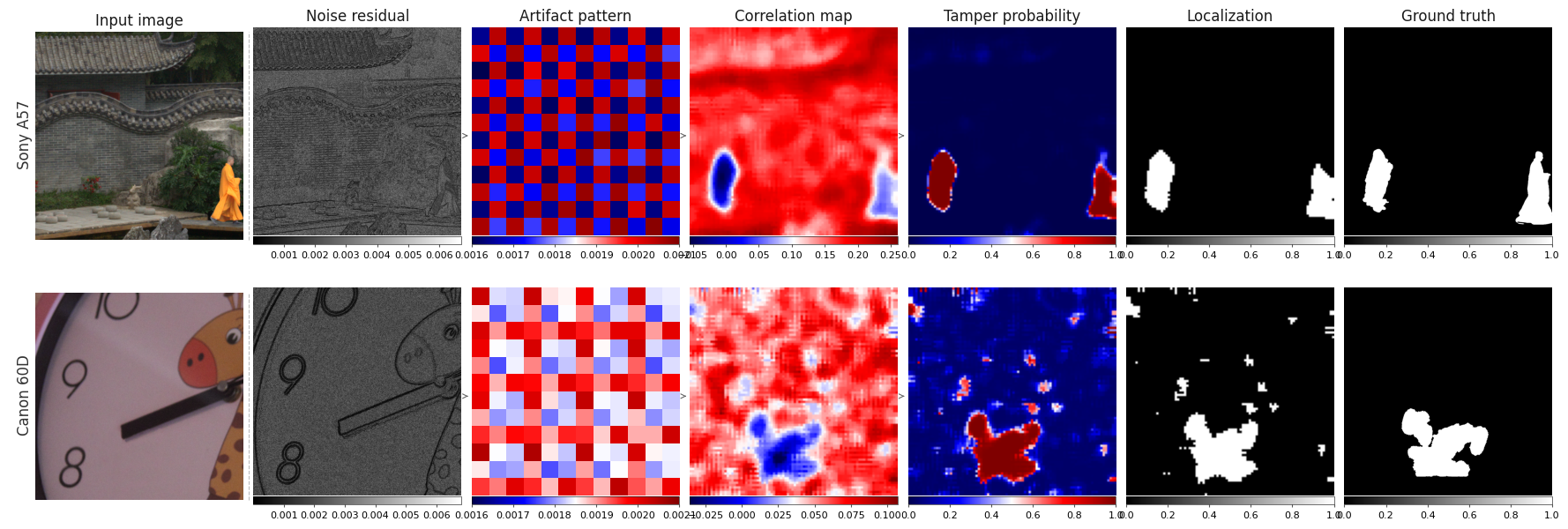}
    \caption{Pipeline execution on two representative examples from the RTD dataset. Each column shows, from left to right: the input image, the noise residual $\mathbf{W}$, the estimated reference pattern  $\mathbf{P}^*$ (for the green channel), the block correlation map $\mathbf{F}$, the soft tampering probability map $\mathbf{T}$, the binary localization mask $\mathbf{S}$, and the ground-truth annotation.}
    \label{fig:pipeline_example}
\end{figure}

In this section, we evaluate the proposed technique employing appropriate metrics for both the tamper heatmap $\mathbf{T}$ and the segmented result $\mathbf{S}$. For this purpose, we employ the well-known Realistic Tampering Dataset (RTD)~\cite{KorusTIFS:2016}. Figure~\ref{fig:pipeline_example} illustrates representative pipeline outputs on two images captured with a Sony $\alpha$57 and a Canon 60D, respectively. It is worth noting that, besides the expected $2\times2$ CFA periodic pattern in the former, the latter shows an unexpected 3$\times$3 pattern that allowed to accurately determine the tampered region. 

\subsection{Evaluation}
\label{subsec:eval_protocol}

Rigorous quantitative evaluation of image manipulation localization methods requires metrics that capture complementary aspects of detection quality at the pixel level. We adopt the following set of measures, all computed at pixel granularity against the ground-truth binary mask.% $\mathbf{S} \in \{0,1\}^{M \times N}$.

\subsubsection{Pixel-level classification metrics.}
Precision, recall, and F$_1$-score assess the quality of the binary localization mask produced by thresholding the soft map $\mathbf{T}$. Precision measures the fraction of pixels flagged as tampered that are genuinely manipulated, penalizing false alarms; recall measures the fraction of truly tampered pixels that are correctly detected, penalizing missed regions. The F$_1$-score is their harmonic mean and provides a single balanced indicator. Pixel accuracy is reported for completeness but is of limited diagnostic value when the tampered region is small relative to the image, as a trivial all-authentic predictor can achieve high accuracy.

\subsubsection{Matthews Correlation Coefficient (MCC).}
MCC is a balanced measure of binary classification quality that
accounts for all four entries of the confusion matrix:
\begin{equation}
    \mathrm{MCC} = \frac{\mathrm{TP}\cdot\mathrm{TN} -
    \mathrm{FP}\cdot\mathrm{FN}}
    {\sqrt{(\mathrm{TP}+\mathrm{FP})(\mathrm{TP}+\mathrm{FN})
    (\mathrm{TN}+\mathrm{FP})(\mathrm{TN}+\mathrm{FN})}},
    \label{eq:mcc}
\end{equation}
\noindent and ranges from $-1$ (total disagreement) to $+1$ (perfect prediction), with $0$ corresponding to random labeling. MCC is particularly robust to class imbalance and is therefore preferable to accuracy and F$_1$-score as a primary indicator when the authentic and tampered pixel populations are of very different sizes~\cite{Chicco:2020}.

\subsubsection{Area under the ROC curve (AUC-ROC).} The AUC-ROC is computed from the soft tampering probability map $\mathbf{T}$ prior to thresholding and measures the method's intrinsic discriminative ability independently of any threshold choice. An AUC close to unity indicates that the scoring function assigns consistently higher values to tampered pixels than to authentic ones across all possible operating points.

Pixel-wise TP, TN, FP, and FN counts are retained internally for computing the above metrics but are not reported in the summary tables. The binary masks required for IoU and MCC are obtained via Otsu's thresholding of $\mathbf{T}$ (Section~\ref{subsec:gmm_scoring}), and results are averaged over all images in each dataset.

\subsection{Quantitative Results}
\label{subsec:results}
\begin{figure}
    \centering
    \includegraphics[width=\linewidth]{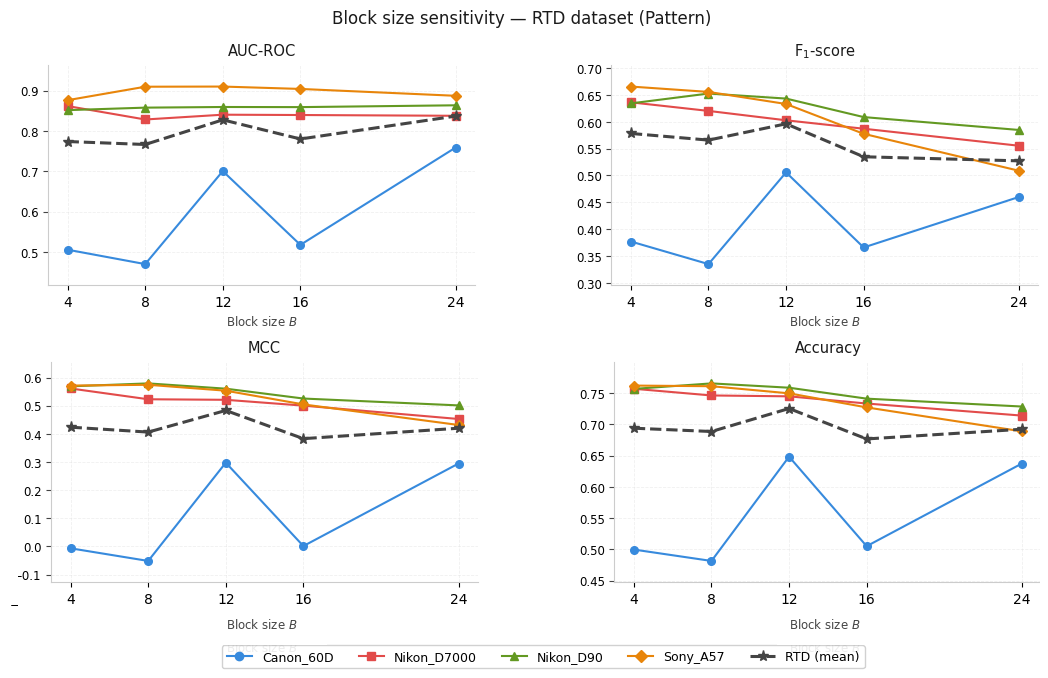}
    \caption{Effect of block size $B$ on pixel-level localization performance across camera models of the RTD dataset. The dashed line denotes the per-metric average over all devices.}
    \label{fig:block_size_sensitivity}
\end{figure}

Figure~\ref{fig:block_size_sensitivity} reports the effect of block size $B$ on localization performance across the four camera models of the RTD dataset. Three cameras --- Nikon D7000, Nikon D90, and Sony A57 --- produce consistent results across all values of $B$, with AUC-ROC above $0.85$, F$_1$-score in the range $0.55$--$0.67$,
and MCC between $0.43$ and $0.59$, indicating that the pipeline is robust to block size for these devices. The RTD dataset mean (dashed) follows the same stable trend, remaining above $0.82$ in AUC-ROC and $0.40$ in MCC for all tested values.

The Canon 60D constitutes a clear outlier: all four metrics exhibit high variance across block sizes, with MCC oscillating between $-0.09$ and $0.30$ and AUC-ROC ranging from $0.47$ to $0.70$. This instability suggests that the periodic CFA artifact pattern of the Canon 60D is not reliably captured at any single block size under the current pipeline configuration, and warrants further investigation into device-specific demosaicing characteristics.

Across all devices, performance tends to be slightly higher at $B \in \{8, 12\}$ than at the extremes $B = 4$ and $B = 24$. Small blocks provide insufficient support for a reliable correlation estimate, while large blocks reduce spatial resolution and may straddle authentic and tampered regions simultaneously. 

Table~\ref{tab:sota_comparison} compares the proposed method against three passive localization approaches on the RTD dataset. The proposed method achieves the highest precision ($0.825$), AUC ($0.839$), IoU ($0.725$), and MCC ($0.483$), indicating superior discriminative ability and spatial localization accuracy overall. The gain in AUC over the next best competitor~\cite{Gonzalez:2023} ($0.781$) confirms that the soft tampering map $\mathbf{T}$ ranks tampered pixels more reliably across all operating thresholds.

It is worth noting that previous methods exhibit reduced performance on the Canon 60D subset, whose proprietary image processing implementation produces atypical artifact patterns that are not well captured at any single block size under the current configuration (Section~\ref{subsec:results}), except for the proposed technique. Because this subset is included in the dataset average, the aggregate metrics reported in Table~\ref{tab:sota_comparison} underestimate the true localization accuracy on well-behaved devices. In this sense, the performance gap between the proposed method and competing approaches may be closer, suggesting that the principal advantage of the proposed pipeline lies in its flexibility: by estimating the reference artifact pattern directly from the suspect image and selecting between signed and absolute residual forms via DCT-based periodicity analysis, the method adapts to a broader range of artifact configurations than methods relying on fixed filter assumptions.

Park~\cite{Park:2021} achieves the highest recall and F$_1$-score driven by a more aggressive localization strategy, but this comes at the expense of precision ($0.689$) and IoU ($0.155$), indicating a possible substantial over-segmentation. Ferrara~\cite{Ferrara:2012} performs consistently below the remaining methods across all metrics, reflecting the limitations of earlier handcrafted approaches on the device diversity present in the RTD dataset.

% ════════════════════════════════════════════════════════════════════════════
% TABLE 3 — Comparison with state-of-the-art methods
% ════════════════════════════════════════════════════════════════════════════
\begin{table}[!h]
\renewcommand{\arraystretch}{1.25}
\caption{Comparison with blind image manipulation localization methods. Results are averaged over all datasets. Best result per metric in bold.}
\label{tab:sota_comparison}
\centering
\begin{tabular}{lcccccc}
\toprule
\multirow{2}{*}{\textbf{Method}} &
\multicolumn{4}{c}{\textbf{Metric}} \\
\cmidrule(lr){2-7} &
\textbf{Precision} &
\textbf{Recall} &
\textbf{F\textsubscript{1}} &
\textbf{IoU} &
\textbf{MCC} &
\textbf{AUC} \\
\midrule
Ferrara~\cite{Ferrara:2012}   0.578& 0.445& 0.467& 0.672& 0.163& 0.354& 0.685\\
Gonzalez~\cite{Gonzalez:2023} 0.772& 0.673& 0.670& 0.710& 0.253& 0.450& 0.781\\
Park~\cite{Park:2021}         0.689& \textbf{0.739}& \textbf{0.697}& 0.698& 0.155& 0.417& 0.774\\
\midrule
Proposed                      \textbf{0.825}& 0.526& 0.597& \textbf{0.725}& \textbf{0.329}& \textbf{0.483}& \textbf{0.839}\\
\bottomrule
\end{tabular}
\end{table}

\section{Conclusions and Future Work}
\label{sec:conclusions}

This work presented a zero-shot, training-free passive pipeline for blind image manipulation localization based on the analysis of acquisition artifact patterns embedded in the noise residual of a single suspect image. The method requires no reference image, device fingerprint database, or labeled training data and operates directly from the full-color noise residual without assuming a fixed CFA configuration or block period. The pipeline integrates a principled denoiser selection criterion based on the acquired-to-interpolated noise variance ratio, a channel-aware artifact pattern estimation stage with DCT-based periodicity selection, block-level Pearson correlation analysis, and a two-component Gaussian Mixture Model scoring stage that produces a pixel-level tampering probability map.

Experimental evaluation on the RTD dataset demonstrated competitive localization performance against three state-of-the-art passive methods, with the proposed approach achieving the highest precision, AUC, IoU, and MCC. The ablation study confirmed that a block size of $B = 12$ provides the best balance between correlation reliability and spatial resolution across most camera models, and identified the Canon 60D as a challenging case whose device-specific demosaicing characteristics warrant further investigation. The consistently strong performance of the pipeline on Nikon and Sony devices suggests that the artifact pattern estimation procedure effectively isolates the CFA fingerprint for a broad class of demosaicing implementations.

Several directions are identified for future work. First, evaluation on a wider range of forensic benchmarks, including CASIA v2, Columbia, COVERAGE, and IMD2020, is currently in progress and will provide a more comprehensive assessment of generalization across manipulation types, compression levels, and sensor diversity. Second, comparisons against additional methods operating under the same zero-shot constraint, including noise-based and PRNU-free approaches, are planned to situate the proposed pipeline more precisely within the passive forensics landscape. Third, the failure mode observed for the Canon 60D motivates an investigation into adaptive block size selection strategies that could accommodate device-specific periodicity, potentially replacing the fixed candidate set $\mathcal{B}$ with a data-driven search guided by the DCT energy criterion of Eq.~(\ref{eq:pattern_selection}).

Finally, extension of the framework to handle AI-generated and diffusion-based manipulations represents a significant open challenge: unlike conventional splicing or copy--move operations, generative edits do not disrupt CFA interpolation traces in the residual in a predictable manner, as the synthesized content may exhibit its own coherent noise structure. Addressing this requires either the incorporation of complementary forensic signals (such as frequency-domain inconsistencies or semantic boundary artifacts) or the development of residual analysis techniques capable of distinguishing camera-native noise from generative model noise, both of which constitute promising directions for subsequent work.

%\section{Experiments and Results}
%\input{document/5_results}

%\section{Conclusions and future work}
%\input{document/6_discussion}

\bibliographystyle{splncs04}
\bibliography{biblio.bib}

\end{document}